\documentclass[letterpaper, 10pt, conference]{ieeeconf}  

\IEEEoverridecommandlockouts    
\pdfoutput=1

\usepackage{graphics}           
\usepackage{epsfig}             
\usepackage{times}              
\usepackage{amsmath}            
\usepackage{amssymb}            
\usepackage{balance}            
\usepackage{cite}               
\usepackage{graphicx}           
\usepackage{physics}            
\usepackage{xcolor}             
\usepackage{tikz}               

\usepackage{hyperref}           
\usepackage{caption}            
\usepackage{subcaption}         
\usepackage{lipsum}             
\usepackage{booktabs}           
\usepackage{tabularx}           
\usepackage{wrapfig}            
\usepackage{multirow}           
\usepackage[section]{placeins}  
\usepackage{siunitx}            
\usepackage{pifont}

\usepackage{booktabs}
\usepackage{makecell}
\usepackage[nointegrals]{wasysym}   

\newcommand{\sfull}{\CIRCLE}        
\newcommand{\spart}{\LEFTcircle}    
\newcommand{\snone}{\Circle}        

\title{\LARGE \bf
From Gameplay to Policy: Towards Scalable Robot Data Collection via Gamified Robot-Free Interaction
}

\author{
Zheng Li$^{1*}$, Liang Zhu$^{1*}$, Junzhe Wang$^{1*}$, Huayuan Chen$^{1}$, Ziyun Liu$^{1}$, Jiahang Cao$^{2}$, \\
Xinyu Sheng$^{1}$, Pei Qu$^{1}$, Yufei Jia$^{3}$, Ximeng Zhang$^{1}$, Jiarui Xie$^{1}$, Zizhao Yuan$^{1}$,\\
Haoang Li$^{1}$, Yi Cai$^{1}$, Jinni Zhou$^{1\dag}$, Jun Ma$^{1}$%
\thanks{*Equal contribution, \dag Corresponding Author.}
\thanks{$^{1}$The Hong Kong University of Science and Technology (Guangzhou).
        {\tt\small \{zli514, lzhu686, jwang787, hchen758, zliu176, xsheng420, pqu458, xzhang305, jxie778, zyuan521\}@connect.hkust-gz.edu.cn, 
        \{haoangli, yicai, eejinni, eejma\}@hkust-gz.edu.cn}}%
\thanks{$^{2}$The University of Hong Kong.
        {\tt\small jiahang@connect.hku.hk}}
\thanks{$^{3}$Tsinghua University.
        {\tt\small jyf23@mails.tsinghua.edu.cn}}
}

\begin{document}
\maketitle

\begin{abstract}

Learning generalizable robot manipulation policies requires large-scale and diverse interaction data, yet collecting real-world demonstrations remains costly and difficult to scale. Existing approaches to data collection are either dependent on specific robot hardware that limits crowdsourcing and transferability, or suffer from incomplete annotation and limited behavioral diversity. Inspired by how games sustain long-term human engagement, we explore an alternative paradigm that turns data collection into an engaging gameplay experience and transfers the resulting human manipulation experience to real robots. We present Project Kitchen, a VR-based gamified egocentric data collection platform that elicits diverse, goal-directed manipulation while remaining independent of specific robot embodiments and hardware, making it applicable to broader and potentially large-scale deployment. To bridge the game-to-real gap, we further introduce Game2Policy, which extracts embodiment-invariant affordance cues, including contact points and sub-goal states, from gameplay trajectories. An affordance model is pre-trained on game-collected data and then jointly fine-tuned with downstream policies using only a handful of real-robot demonstrations. Experiments show that Game2Policy improves average success rates by 10.0 points in simulation and 18.3 points on real robots in the few-shot setting. User studies and quantitative analyses further show that Project Kitchen promotes diverse manipulation behaviors and provides an engaging data collection experience. These results demonstrate the potential of gamified virtual environments as a scalable source of manipulation knowledge. The platform and code will be released upon acceptance.


\end{abstract}

\section{Introduction} \label{sec 1}

Over the past decade, data-driven approaches, especially imitation learning~\cite{2025dp, ACT} and reinforcement learning~\cite{newbury2023deep, li2026manivid}, have reshaped robotics by enabling robots to acquire complex sensorimotor skills from experience, such as dexterous~\cite{luo2025precisedex2}, bimanual~\cite{qu2026omnidp}, and contact-rich~\cite{yu2026forcevla} manipulation. These advances, similar to those in computer vision and NLP, promise strong generalization through large datasets and could greatly advance automation. However, collecting robot manipulation data is far less convenient than scraping web text, as it demands constant human supervision for resets, safety, and demonstrations, making the process slow, expensive, and hard to scale~\cite{ohkawa2026yubi, ye2026data}. This severely limits the deployment of generalist robotic policies.

Existing approaches to robot data collection suffer from several inherent limitations. Teleoperation with real hardware is slow and embodiment-specific, which severely constrains both scalability and transferability~\cite{ACT, wu2024gello}. Simulated data can be produced in bulk, but the trajectories it generates are often too homogeneous,
failing to capture the richness of real-world human manipulation
~\cite{dong2026hypersim, zhang2026affordgen}. Specialized systems such as UMI depend on custom gripper designs, rendering crowdsourcing impractical~\cite{chi2024universal}. Human videos contain rich motion information but lack action labels and require extensive manual filtering to extract useful clips~\cite{feng2026human}.
An interesting analogy reveals an alternative possibility: when placed in immersive, goal-oriented settings such as games, humans tend to engage persistently with a slower onset of fatigue, and spontaneously generate diverse interactions rather than repetitive demonstrations. 
This raises the question: can we leverage video game incentives to make data collection both engaging and scalable?


\begin{figure}[!t] 
\vspace{5pt}
\raggedleft 
\includegraphics[width=0.48\textwidth]{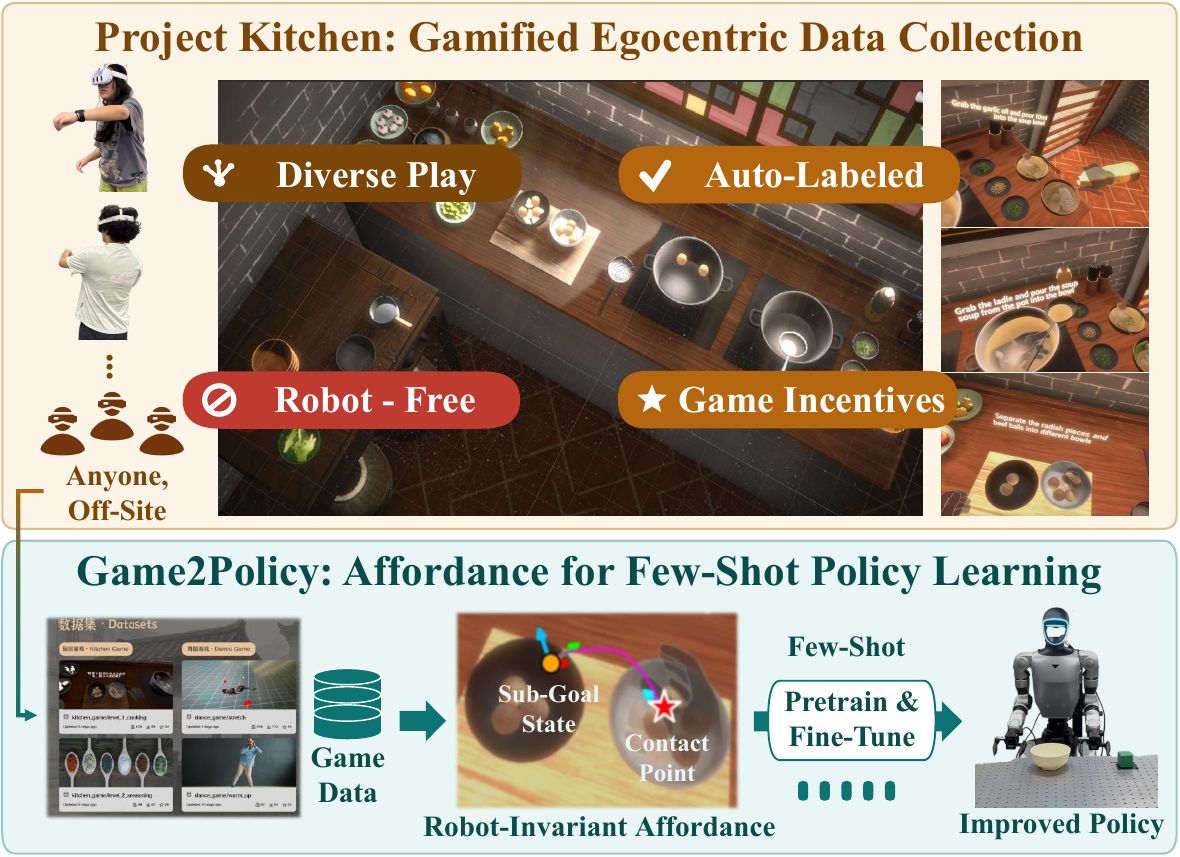}
\caption{\textbf{From gameplay to policy.} Our system collects diverse human manipulation data through VR gameplay (top) and transfers embodiment-invariant affordance cues to real robots with only a handful of demonstrations (bottom).}
\label{fig:teaser}
\end{figure}

\begin{table*}[t]
  \vspace{5pt}
  \centering
  \caption{%
    \textbf{Comparison of data-collection paradigms.}
    \sfull~= supported, \spart~= partially supported,
    \snone~= not supported.
    Our gamified paradigm decouples collection from robot hardware, laboratory settings, and expert operators, lowering the barrier to crowdsourcing, while still providing ground-truth actions, automatic labels, and diverse human manipulation behaviors.
    }
  \label{tab:collection_comparison}
  \footnotesize
  \setlength{\tabcolsep}{6pt}
  \renewcommand{\arraystretch}{1.15}
  \begin{tabular}{@{}c| c| ccc cc cc c@{}}
    \toprule
    & \textbf{Platform}
    & \multicolumn{3}{c}{\textbf{Deployment}}
    & \multicolumn{2}{c}{\textbf{Human Scaling}}
    & \multicolumn{2}{c}{\textbf{Supervision}}
    & \textbf{Data} \\
    \cmidrule(lr){2-2} \cmidrule(lr){3-5} \cmidrule(lr){6-7}
    \cmidrule(lr){8-9} \cmidrule(lr){10-10}
    \textbf{Collection Paradigm}
    & \makecell[c]{Capture Device}
    & \makecell{Robot-\\Free}
    & \makecell{Off-\\Site}
    & \makecell{Risk-\\Free}
    & \makecell{Crowdsourcing-\\Ready}
    & \makecell{Non-\\Expert}
    & \makecell{GT\\Actions}
    & \makecell{Label-\\Free}
    & \makecell{Behavior\\Diversity} \\
    \specialrule{\lightrulewidth}{\aboverulesep}{0pt}
    \specialrule{\lightrulewidth}{\doublerulesep}{\belowrulesep}
    RoboCade~\cite{mirchandani2025robocade}
      & \makecell[c]{3D-printed leader arm}
      & \snone & \sfull & \snone & \spart & \spart  & \sfull & -- & \spart \\
    RoboCrowd~\cite{mirchandani2025robocrowd}
      & \makecell[c]{On-site ALOHA arms}
      & \snone & \snone & \snone & \spart & \spart  & \sfull & \snone & \spart \\
    Human Video~\cite{damen2022epic, grauman2022ego4d}
      & \makecell[c]{Consumer cameras}
      & \sfull & \sfull & \sfull & \sfull & \sfull  & \snone & \snone & \sfull \\
    Real-Robot Teleop.~\cite{khazatsky2024droid}
      & \makecell[c]{Franka + 3 cameras}
      & \snone & \snone & \snone & \snone & \snone  & \sfull & \spart & \spart \\
    UMI~\cite{chi2024universal}
      & \makecell[c]{Handheld gripper}
      & \sfull & \sfull & \sfull & \spart & \spart  & \spart & \spart & \spart \\
    Simulation~\cite{2012mujoco, gao2026nvidia}
      & \makecell[c]{Simulator}
      & \sfull & \sfull & \sfull & \sfull & \spart  & \sfull & \sfull & \snone \\
    \midrule
    \textbf{Ours}
      & \makecell[c]{Consumer VR headset}
      & \sfull & \sfull & \sfull & \sfull & \sfull  & \sfull & \sfull & \sfull \\
    \bottomrule
  \end{tabular}
\end{table*}

Video games are inherently engaging and designed to maintain motivation through clear objectives, immediate feedback, and progressive challenges. Despite gamification's proven value in crowdsourcing applications ~\cite{bamford2009galaxy, shingjergji2022face,  mirchandani2025robocade}, its deliberate application to robot policy learning remains surprisingly rare. The few existing gamification attempts, while effective at boosting user engagement, remain grounded in conventional teleoperation interfaces with superficial reward overlays and are inherently tethered to physical robot hardware. Such reliance limits the number of users who can participate concurrently, complicates deployment beyond laboratory settings, and ultimately hinders large-scale crowdsourced data collection~\cite{mirchandani2025robocrowd, mirchandani2025robocade}.

To address these challenges, we introduce \textbf{Project Kitchen}, a VR-based gamified platform for collecting diverse egocentric manipulation data, and \textbf{Game2Policy}, a learning framework that transfers affordance cues from the collected gameplay to few-shot robot policy learning.
Unlike prior gamification attempts, our system is embodiment-agnostic and decoupled from specific robot hardware, 
making it applicable to crowdsourcing and potentially scalable data collection at low cost. It encourages diverse manipulation through game incentives and enables automatic annotation via subtask success predicates. A high-level comparison with existing manipulation data collection methods is provided in TABLE~\ref{tab:collection_comparison}. From the collected trajectories, we extract transferable, embodiment‑invariant affordance cues (e.g., contact points and sub‑goal states), pre‑train a lightweight affordance prediction model on game data, and jointly fine‑tune it with downstream policies using a handful of real‑robot demonstrations. Comprehensive experiments show average success-rate improvements of 10.0 points in simulation and 18.3 points on real robots in the few-shot setting, effectively reducing reliance on costly demonstrations while preserving the diversity of natural human manipulation. A user study and quantitative metrics further demonstrate the advantages of our data collection paradigm over alternative approaches.

In summary, our core contributions are as follows:

\begin{itemize}

\item We present Project Kitchen, a VR-based gamified egocentric data collection platform that decouples manipulation data collection from robot hardware, lowering barriers to crowdsourced data collection, while using game incentives to elicit diverse manipulation and subtask success predicates to enable automatic annotation.
\item We propose Game2Policy, a framework that extracts embodiment-invariant affordance cues from gameplay trajectories and transfers them to downstream policies via pre-training on game-collected data and joint fine-tuning with a handful of real-robot demonstrations.
\item We conduct comprehensive experiments in simulation and on real robots, showing that our approach consistently improves few-shot policy success rates, and we further demonstrate the advantages of our data collection paradigm over alternatives via a user study and a series of quantitative metrics.

\end{itemize}

\section{Related Work}\label{sec 2}

\subsection{Data Collection Methods for Robot Learning}



Teleoperation remains the predominant approach for collecting real-world robot demonstrations. Systems such as ALOHA~\cite{ACT}, DROID~\cite{khazatsky2024droid} and SONIC~\cite{Luo2026SONIC} enable human operators to directly control robot arms through leader-follower interfaces, generating high-fidelity trajectories with rich physical interactions. The Universal Manipulation Interface (UMI)~\cite{chi2024universal} offers a complementary paradigm by decoupling data collection from specific robot hardware: users collect visual and pose data with handheld grippers, which can be deployed across multiple robot platforms. However, teleoperation is slow, hardware‑dependent, and labor‑intensive, demanding sustained attention from skilled operators, while UMI relies on custom handheld devices that limit crowdsourcing at scale.

Simulation and human videos provide hardware-free alternatives. Physics engines such as MuJoCo~\cite{2012mujoco} and Isaac Sim~\cite{gao2026nvidia} generate demonstrations at scale without physical robot access, while internet-scale human videos offer a vast and naturally diverse corpus of behaviors, with methods like ViViDex~\cite{chen2025vividex} and ConLA~\cite{dai2026conla} learning robotic policies and skills directly from video data. However, simulation-generated trajectories tend to be overly homogeneous and lack the behavioral diversity inherent in human manipulation, which constrains the generalization of imitation learning policies~\cite{jia2024towards}. Human videos, by contrast, lack ground-truth action labels and require extensive manual filtering to extract usable segments for policy learning.


\begin{figure*}[!t] 
\vspace{5pt}
\raggedleft 
\includegraphics[width=\textwidth]{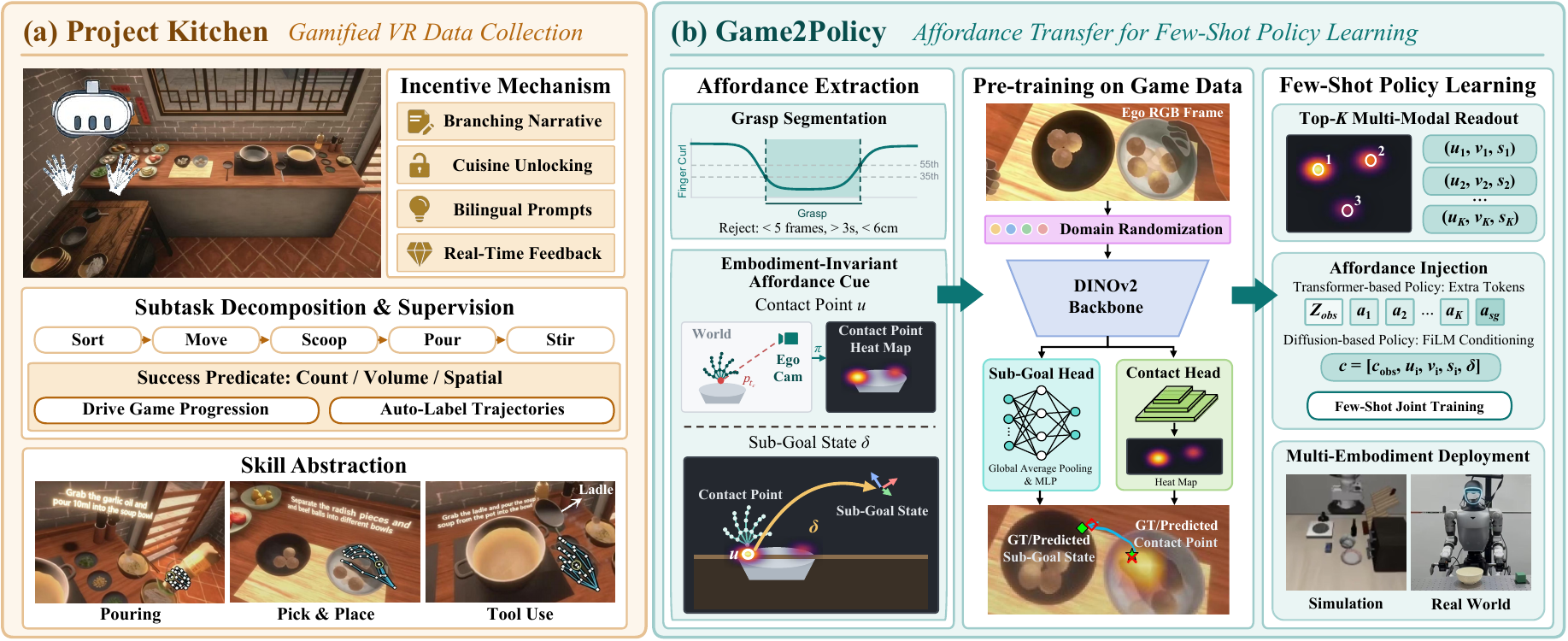}
\caption{\textbf{System overview.} Our system consists of two stages: (a) Project Kitchen, a VR-based gamified platform where users complete cooking tasks via bare-hand interactions, with gameplay-driven automatic segmentation and annotation; (b) Game2Policy, which extracts embodiment-invariant affordance cues (contact points and sub-goals) from gameplay trajectories, pre-trains a lightweight affordance model on game data, and jointly fine-tunes it with downstream policies using only a few demonstrations for deployment.}
\label{fig:system}
\end{figure*}

\subsection{Gamification in Machine Learning}

Gamification has proven effective at sustaining human motivation across diverse domains, from citizen science platforms like Galaxy Zoo~\cite{bamford2009galaxy} to gamified educational tools that improve knowledge retention and skill acquisition~\cite{manninen2025effect}. These successes suggest that well-designed incentives can transform tedious work into engaging play. Recent work has extended gamification to data collection in machine learning, including comprehensive reviews of gamification techniques~\cite{khakpour2021gamereview}, gamified approaches for annotated facial expression data~\cite{shingjergji2022face}, and crowdsourcing of affective communication data through speech emotion recognition~\cite{siamtanidou2025speech}. However, gamification in robot policy learning remains scarce. Existing attempts, though capable of increasing user engagement, repurpose conventional teleoperation with superficial reward scoring, an approach inherently tied to physical hardware~\cite{mirchandani2025robocrowd, mirchandani2025robocade}. This reliance on real-robot interfaces limits large-scale crowdsourcing, narrows behavioral diversity, and creates scalability challenges.

\section{Method}\label{sec:method}

\subsection{System Overview}
Our system turns manipulation data collection into gameplay and transfers the resulting experience to real robots via embodiment-invariant affordance cues. As illustrated in Fig.~\ref{fig:system}, it consists of two stages. (a) Project Kitchen (Sec.~\ref{project_kitchen}) is a VR cooking game where users complete long-horizon recipes via bare-hand interactions. Game incentives are designed to sustain engagement and encourage diverse manipulation trajectories as a by-product of play; recipe subtasks with success predicates automatically annotate the stream, and the resulting clips are organized into three reusable primitives: pick-and-place, pouring, and tool use. (b) Game2Policy (Sec.~\ref{game2policy}) bridges the game-to-real gap without action-level supervision. From gameplay, we extract contact points and sub-goals as embodiment-invariant affordance cues, pre-train a lightweight model on game data with domain randomization, and jointly fine-tune it with only a few real-robot demonstrations for downstream deployment.

\subsection{Project Kitchen: A Gamified Platform for Egocentric Demonstration Collection}\label{project_kitchen}
\textbf{Gamified VR platform and data collection.}
We develop \textit{Project Kitchen}, a VR cooking game for collecting embodiment-agnostic human manipulation demonstrations. Users follow recipe instructions to complete long-horizon cooking tasks, where each recipe consists of an ordered sequence of subtasks involving interactions with ingredients, containers, and utensils. By embedding these subtasks into goal-oriented gameplay with task progression and feedback, manipulation trajectories are collected naturally as a by-product of completing the game rather than through repetitive isolated demonstrations. To provide gameplay incentives that sustain user engagement, the game incorporates region-locked cuisine unlocking, a branching narrative, bilingual prompts, and real-time feedback. The recipe structure further provides semantic organization for the continuous gameplay trajectory, which is later exploited for automatic segmentation and supervision.

The platform is implemented in Unity 2022.3 with hand interaction supported by the Meta XR SDK. During gameplay, we synchronously record 26-joint hand poses, object poses and task-relevant states, grasp/release events, and task progress at 30~Hz, forming the raw demonstration stream for subsequent robot-oriented processing.

\textbf{Gameplay-driven segmentation and supervision.}
Given the continuous gameplay stream, we exploit the game logic itself to automatically identify and annotate individual manipulation segments. Each recipe subtask is associated with a success predicate evaluated from the current scene state, including spatial-, count-, and volume-based conditions. Once a predicate is satisfied, the corresponding frame provides both the subtask boundary and its success label, while grasp/release events offer additional temporal anchors for interaction-centric segmentation. For continuous state-changing interactions such as pouring, task states are represented in physically interpretable quantities. Specifically, a normalized fill level $\ell \in [0,1]$ is converted to absolute volume as $V=\ell C$, where $C$ is the vessel capacity, enabling predicates such as required transferred volume to be evaluated consistently across containers. A lightweight volume-conserving abstraction is sufficient for this purpose, avoiding unnecessary fluid simulation while retaining task-relevant supervision.

\textbf{Robot-oriented skill abstraction.}
After segmentation, the resulting gameplay clips are further organized according to their underlying manipulation semantics rather than their specific cooking context. In this work, we focus on three representative and reusable manipulation primitives:

\textit{(i) Pick-and-place.}
This primitive captures spatial object relocation and consists of reaching, grasp acquisition, object transport, spatial placement, and release. It represents a fundamental manipulation pattern in which the primary objective is to move an object from its initial configuration to a target region while maintaining stable grasp control.

\textit{(ii) Pour.}
This primitive extends object transport with continuous orientation control and regulation of the resulting object-state change. In addition to grasping and repositioning the container, the operator must coordinate its pose and tilt angle to control material transfer, making pouring representative of continuous state-changing manipulation.

\textit{(iii) Tool use.} 
This primitive captures tool-mediated interaction, involving grasping, tool--object alignment, constrained contact motion, and material transfer. Unlike direct manipulation, tool use such as ladling requires coordinated control of both tool and target, providing demonstrations of contact-rich manipulation.


These primitives are not constructed to match particular downstream benchmark tasks. Instead, they are abstracted from manipulation behaviors that naturally arise during gameplay and serve as broadly useful units for subsequent robot learning.

\subsection{Game2Policy: Affordance Transfer for Few-Shot Policy Learning}\label{game2policy}

To mitigate the gap between gameplay and robot execution, inspired by prior works~\cite{xu2025a0, wang2023mimicplay}, we leave action-level supervision from game data to future work and instead extract transferable information from trajectories, specifically contact points and sub-goal states that mark task progress. These affordance cues are embodiment-invariant and serve as conditioning signals for downstream policy learning. We pre-train an affordance prediction model on game data, and then jointly fine-tune it with the downstream policy model on a small set of demonstrations via few-shot learning. This strategy substantially reduces the need for costly physical robot data collection.

\textbf{Affordance extraction.}
We source supervision from gameplay of \textit{Project Kitchen} with 26 joints per hand tracked. For each grasp $(t_c, t_r)$, the contact point is defined as the palm world position at $t_c$, projected into preceding frames:
\begin{equation}
  \mathbf{u}_t = \pi\big(\mathbf{K},\ \mathbf{T}^{wc}_t\, \mathbf{p}_{\tau(t)}\big),
  \tau(t) = \begin{cases}
  t_c, & t \in [t_c - H_r,\ t_c]  \\
  t,   & t > t_c  
  \end{cases}
\label{eq:contact}
\end{equation}
where $\pi$ is the projection function, $\mathbf{K}$ the camera intrinsics, $\mathbf{T}^{wc}_t$ the world-to-camera extrinsic at frame $t$, $\mathbf{p}_{t_c}$ the palm position at contact, and $H_r = 1.5\,\text{s}$ the lookback horizon, and $\tau(t)$ selects the contact position before $t_c$ and the current position afterwards. Because Eq.~\eqref{eq:contact} uses each frame's own camera parameters, the label remains correctly localized as the player moves: it is anchored in the world rather than the image, requiring no manual annotation. Projections with $z_t \le 0$ or falling outside the image are masked. We supervise a Gaussian heatmap rather than regressing coordinates, as contact locations are inherently multimodal (e.g., either rim of a bowl). The sub-goal encodes a relative pose in the contact frame:
\begin{equation}
\boldsymbol{\delta} = \big[\mathbf{R}_{t_c}^{\top}(\mathbf{p}_{t_r^{\mathrm{kp}}} - \mathbf{p}_{t_c}),\ \rho(\mathbf{R}_{t_c}^{\top}\mathbf{R}_{t_r^{\mathrm{kp}}})\big] \in \mathbb{R}^9,
\end{equation}
where $\mathbf{R}_{t_c}$ is the palm rotation at contact, $t_r^{\mathrm{kp}}$ a retracted endpoint within the grasp interval, and $\rho$ the six-dimensional rotation representation. Expressing $\boldsymbol{\delta}$ relative to the contact frame cancels world frame and camera motion, making it comparable across scenes and embodiments. It is expressed in a human hand frame, not a robot end-effector frame, and is therefore treated as an embodiment-invariant conditioning signal rather than an executable target.

\textbf{Trajectory segmentation.}
Grasp intervals are derived entirely from hand kinematics without manual annotation. Rather than detecting hand-object proximity, which yields many false positives in scenes cluttered with interactable objects, we use finger curl as a direct grasp indicator. Thresholds are set per recording (35th/55th percentiles) to absorb variation across players and object sizes, while hysteresis suppresses chattering near the threshold. Coarse closure onset is refined to the local minimum of palm speed. Intervals under 5 frames are discarded as jitter; those exceeding 3\,s are truncated; those with pre-contact motion under 6\,cm are rejected. The sub-goal endpoint is retracted from the interval end until displacement is plausible and camera motion is minimal, which excludes locomotion where the hand's world displacement is dominated by body motion rather than manipulation.

\begin{table*}[t]
  \vspace{5pt}
  \centering
  \caption{\textbf{Simulation experiment results.} Our method consistently improves success rates across all three base policies and tasks, with the largest gains in the few-shot regime, where game pre-training recovers most of the data-rich performance using one tenth of the target demonstrations.}
  \label{tab:exp2_main}
  \small
  \setlength{\tabcolsep}{5pt}
  \renewcommand{\arraystretch}{1.15}
  \resizebox{\linewidth}{!}{%
  \begin{tabular}{@{}c|c| cccc|cccc@{}}
    \toprule
    & & \multicolumn{4}{c}{\textbf{5 Demos}}
      & \multicolumn{4}{c}{\textbf{50 Demos}} \\
    \cmidrule(lr){3-6} \cmidrule(lr){7-10}
    \textbf{Base Policy} & \textbf{Method}
    & Bottle-on-Cabinet & Cheese-in-Bowl & Bowl-on-Plate & Avg.
    & Bottle-on-Cabinet & Cheese-in-Bowl & Bowl-on-Plate & Avg. \\
    \specialrule{\lightrulewidth}{\aboverulesep}{0pt}
    \specialrule{\lightrulewidth}{\doublerulesep}{\belowrulesep}
    \multirow{3}{*}{ACT~\cite{ACT}}
      & Target-Only      & $69.6 \pm 6.2$ & $54.4 \pm 1.7$ & $72.4 \pm 7.1$ & $65.5$
                     & $\mathbf{85.2 \pm 1.8}$ & $60.4 \pm 5.7$ & $81.2 \pm 3.0$ & $75.6$ \\
      & Scratch  & $66.8 \pm 4.1$ & $56.0 \pm 5.5$ & $76.0 \pm 4.5$ & $66.3$
                     & $77.6 \pm 3.3$ & $64.4 \pm 5.5$ & $79.2 \pm 8.3$ & $73.3$ \\
      & \textbf{Ours}& $\mathbf{80.0 \pm 6.3}$ & $\mathbf{67.6 \pm 3.8}$ & $\mathbf{81.6 \pm 2.6}$ & $\mathbf{76.4}$
                     & $83.2 \pm 6.4$ & $\mathbf{69.2 \pm 5.8}$ & $\mathbf{84.8 \pm 2.7}$ & $\mathbf{79.1}$ \\
    \midrule
    \multirow{3}{*}{DP~\cite{2025dp}}
      & Target-Only      & $64.8 \pm 3.0$ & $44.4 \pm 1.5$ & $78.8 \pm 1.6$ & $62.7$
                     & $76.4 \pm 1.5$ & $64.8 \pm 6.0$ & $\mathbf{90.8 \pm 3.7}$ & $77.3$ \\
      & Scratch  & $60.8 \pm 6.8$ & $41.2 \pm 4.1$ & $78.4 \pm 5.1$ & $60.1$
                     & $75.6 \pm 2.9$ & $58.4 \pm 4.1$ & $88.0 \pm 2.2$ & $74.0$ \\
      & \textbf{Ours}& $\mathbf{73.6 \pm 5.6}$ & $\mathbf{53.6 \pm 4.1}$ & $\mathbf{85.6 \pm 2.9}$ & $\mathbf{70.9}$
                     & $\mathbf{82.0 \pm 2.8}$ & $\mathbf{67.6 \pm 3.4}$ & $89.2 \pm 3.5$ & $\mathbf{79.6}$ \\
    \midrule
    \multirow{3}{*}{SmolVLA~\cite{shukor2025smolvla}}
      & Target-Only      & $67.6 \pm 8.6$ & $52.0 \pm 4.5$ & $78.8 \pm 4.6$ & $66.1$
                     & $86.4 \pm 3.8$ & $58.8 \pm 5.2$ & $84.4 \pm 4.6$ & $76.5$ \\
      & Scratch  & $72.8 \pm 3.3$ & $51.2 \pm 4.6$ & $80.4 \pm 4.6$ & $68.1$
                     & $\mathbf{90.0 \pm 2.0}$ & $68.0 \pm 8.2$ & $86.0 \pm 4.7$ & $81.3$ \\
      & \textbf{Ours}& $\mathbf{80.4 \pm 4.6}$ & $\mathbf{66.0 \pm 2.8}$ & $\mathbf{84.4 \pm 4.3}$ & $\mathbf{76.9}$
                     & $87.2 \pm 3.0$ & $\mathbf{71.6 \pm 5.4}$ & $\mathbf{91.6 \pm 3.8}$ & $\mathbf{83.5}$ \\
    \bottomrule
  \end{tabular}
  }
\end{table*}

\textbf{Pre-training the affordance model.}
A lightweight network predicts contact points and sub-goals from a single RGB frame. Features are extracted by a frozen DINOv2 backbone; all trainable parameters are in the prediction heads. The feature map splits into two branches: convolutions produce a $64 \times 64$ contact heatmap, while global pooling feeds an MLP for sub-goal regression. This asymmetry is deliberate: \emph{where to grasp} is a \emph{local} question, so that branch keeps the spatial grid, whereas \emph{where to move next} depends on the \emph{global} scene layout, which pooling retains while discarding position. The contact point is supervised by heatmap classification $\mathcal{L}_{\mathrm{hm}}$ and coordinate regression $\mathcal{L}_{\mathrm{coord}} = \|\hat{\mathbf{u}} - \mathbf{u}\|_2^2$ via soft-argmax; the sub-goal by Euclidean position error and geodesic rotation distance:
\begin{equation}
\mathcal{L} = \mathcal{L}_{\mathrm{hm}} + \mathcal{L}_{\mathrm{coord}} + \|\hat{\mathbf{d}} - \mathbf{d}\|_2^2 + d_{\mathrm{geo}}(\hat{\mathbf{R}}, \mathbf{R}).
\end{equation}
To bridge the visual domain gap, we additionally augment inputs with domain randomization on lighting and background during pre-training. All losses use only valid frames, with episode-level splits. 

\begin{figure}[!t] 
\centering 
\includegraphics[width=0.45\textwidth]{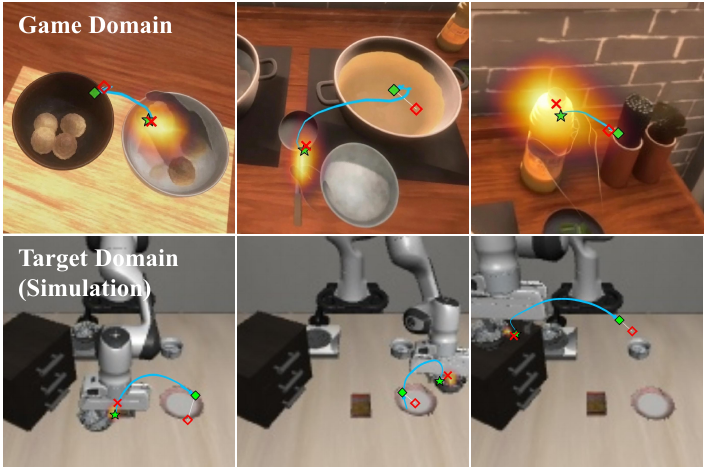}
\caption{\textbf{Snapshot and affordance visualization in the game and simulation domains.} Green stars and diamonds denote ground-truth contact points and sub-goal points, respectively; red markers denote the corresponding predictions.}
\label{fig:simulation}
\end{figure}

\textbf{Downstream injection and few-shot policy training.}
The pre-trained affordance model is queried online alongside the downstream policy at each step. Its differentiable readout lets the behavior-cloning loss back-propagate through it, enabling task adaptation while a learning rate lower than in pre-training preserves pre-trained knowledge. To preserve multi-modality in grasping, we extract the top $K$ heatmap peaks via windowed soft-argmax with non-maximum suppression, each with a response strength $s_i$ that acts as an implicit confidence, allowing the policy to reason over multiple plausible contacts. For transformer policies (ACT, SmolVLA), each candidate contributes $(u_i, v_i, s_i)$, and the sub-goal contributes $\boldsymbol{\delta}$; all are zero-padded to width $d_a$ and appended as tokens:
\begin{equation}
\begin{aligned}
\mathbf{Z} &= [\mathbf{Z}_{\mathrm{obs}},\ \mathbf{a}_1,\dots,\mathbf{a}_K,\ \mathbf{a}_{\mathrm{sg}}],\\
\mathbf{a}_i &= [u_i, v_i, s_i, \mathbf{0}_9],\\
\mathbf{a}_{\mathrm{sg}} &= [\mathbf{0}_3, \boldsymbol{\delta}],
\end{aligned}
\end{equation}
where $\mathbf{Z}_{\mathrm{obs}}$ is the observation token sequence and $\mathbf{0}_n$ denotes $n$-dimensional zero padding. For Diffusion Policy, which uses a single conditioning vector $\mathbf{c}$ via FiLM, affordance is concatenated directly:
\begin{equation}
\mathbf{c} = [\mathbf{c}_{\mathrm{obs}},\ u_1, v_1, s_1,\ \dots,\ u_K, v_K, s_K,\ \boldsymbol{\delta}],
\end{equation}
where $\mathbf{c}_{\mathrm{obs}}$ is the observation conditioning vector. In the downstream setting, each task is provided with very few demonstrations, and this few-shot training approach substantially reduces the need for costly real-robot data collection.

\section{Experiments}

\subsection{Experiment Setup}

We initially collect 10 hours of gameplay data from Project Kitchen and pre-train our affordance model on this dataset before fine-tuning it on downstream tasks.

\textbf{Simulation setup.} We evaluate on three LIBERO-Goal tasks: put the wine bottle on top of the cabinet (\textit{Bottle-on-Cabinet}), put the cream cheese in the bowl (\textit{Cheese-in-Bowl}), and put the bowl on the plate (\textit{Bowl-on-Plate}). Each task is trained with $5$ and $50$ demonstrations to separate few-shot from data-rich regimes. To evaluate whether the transferred affordance generalizes across policy architectures, we instantiate three architecturally distinct base policies: ACT~\cite{ACT}, Diffusion Policy~\cite{2025dp}, and SmolVLA~\cite{shukor2025smolvla}. All policies share identical observation spaces, training schedules, and evaluation protocols, differing only in conditioning and initialization. Results are reported as mean\,$\pm$\,std over $3$ seeds with $50$ rollouts per seed.

\textbf{Real-robot setup.} Experiments run on a Unitree G1 humanoid equipped with an egocentric head camera and a Dex1-1 parallel gripper, which tests whether affordance cues extracted from bare-hand gameplay transfer to a robot end-effector with a fundamentally different morphology. We evaluate the three manipulation primitives abstracted from gameplay: pick-and-place, pour, and tool use. Following the few-shot protocol, each task receives only $10$ teleoperated demonstrations and uses ACT as the base policy, evaluated over $20$ trials per task with random initial placements.

\textbf{Compared methods.} To isolate the contribution of game-collected affordance rather than architectural capacity, we compare against two controls. \textit{Target-Only} trains on target demonstrations alone, without any affordance conditioning or game data. \textit{Scratch} keeps our full affordance-conditioned architecture but randomly initializes the affordance model, learning jointly from target demonstrations only (i.e., no game pre-training). The first control measures the end-to-end benefit of our pipeline, while the second removes the confounding effect of extra conditioning capacity, attributing gains specifically to transferred knowledge.

\textbf{Evaluation metrics.} We report task success rate as the primary metric. Beyond policy performance, we further characterize the collected data along two axes (TABLE~\ref{tab:data_quality}): \textit{Label Quality}---compared against egocentric human video~\cite{damen2022epic} via signal dropouts per second, hand-label flips, coverage, PCK@2 contact localization, and required manual
annotations; and \textit{Behavioral Diversity}---compared against simulation-generated demonstrations via approach spread, number of unimodal groups, occupancy entropy, and contact-chain length. Together, these metrics test whether our platform yields annotation-free labels of teleoperation-grade quality while retaining the behavioral richness that simulated data lacks. We additionally conduct a user study with $10$ participants (Fig.~\ref{fig:user_study}) to assess the engagement and usability of the collection process itself.

\subsection{Simulation Experiments}

\begin{table}[t]
    \vspace{5pt}
    \centering
    \caption{\textbf{Real-world experiment results.} Game pre-training consistently improves success rates across all three primitives in the few-shot setting.}
    \label{tab:real_world}
    \setlength{\tabcolsep}{5pt}
    \resizebox{0.9\linewidth}{!}{
    \begin{tabular}{l|ccc|c}
        \toprule
        \textbf{Method}
        & Pick \& Place
        & Pour
        & Tool Use
        & Total \\
        \midrule

        Target-Only
        & 10/20 & 8/20 & 12/20 & 30/60\\

        Scratch
        & 11/20 & 8/20 & 9/20 & 28/60\\

        \textbf{Ours}
        & \textbf{14/20} & \textbf{12/20} & \textbf{15/20} & \textbf{41/60}\\

        \bottomrule
    \end{tabular}
    }
\end{table}

\begin{figure}[!t] 
\centering 
\includegraphics[width=0.48\textwidth]{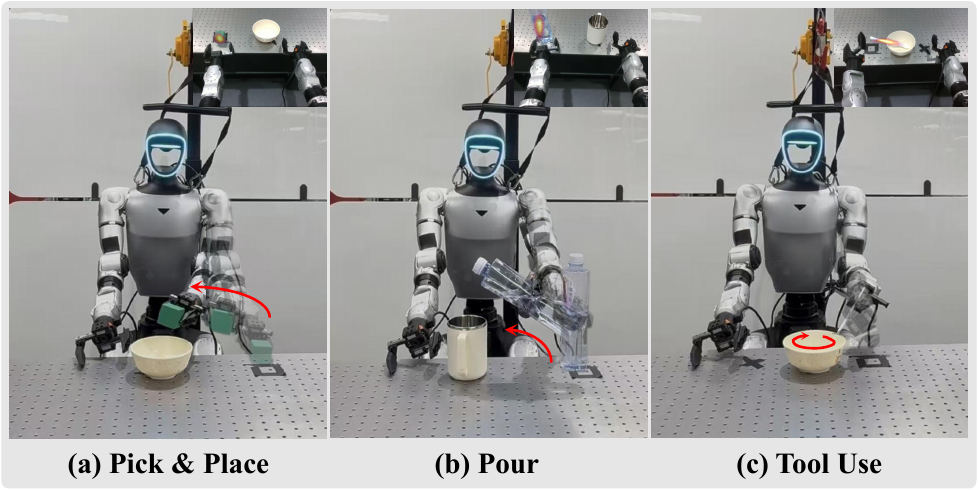}
\caption{\textbf{Real-world experiment snapshots.} We conduct real-world experiments using Unitree G1 with Dex1-1 Gripper.}
\label{fig:real}
\end{figure}

We evaluate Game2Policy on three tasks from LIBERO-Goal benchmark. The game-pretrained affordance model is jointly fine-tuned in the target simulation domain, as illustrated in Fig.~\ref{fig:simulation}, and results are summarized in Table~\ref{tab:exp2_main}. The improvement of Game2Policy stems from knowledge transferred by game pre-training, not from increased architectural capacity. With 5 demonstrations, our method outperforms Target-Only by $10.0$ points on average; with 50 demonstrations the advantage shrinks to $4.3$ points. The most direct evidence comes from the Scratch control: it retains the identical affordance-conditioned architecture but randomly initializes the affordance model, and achieves a similar score as Target-Only. This indicates that merely adding the conditioning branch yields no measurable benefit; what matters is the affordance prior learned during pre-training. Moreover, the improvement appears consistently across three substantially different policy architectures including ACT, DP and SmolVLA, demonstrating that the transferred signal generalizes across diverse policy architectures.

The benefit concentrates where demonstrations are scarcest. With 5 demonstrations our average score of $74.7$ already recovers most of the 50-demonstration Target-Only result ($76.5$) while using one tenth of the target-domain data; when demonstrations grow to 50, the margin narrows from $+10.0$ to $+4.3$, consistent with the affordance acting as a prior that demonstrations eventually supply on their own. 






\subsection{Real-World Experiments}

We deploy Game2Policy to the Unitree G1 with the Dex1-1 gripper on the three primitives abstracted from gameplay, as shown in Fig.~\ref{fig:real}, with results summarized in TABLE~\ref{tab:real_world}. Although the affordance model is pre-trained solely on bare-hand gameplay, conditioning ACT on its predictions raises the overall success rate from 30/60 to 41/60 in the few-shot setting, with consistent gains on all three primitives. The Scratch achieves comparable performance to Target-Only (28/60), which again attributes the gain to game pre-training rather than added conditioning capacity. Despite the large morphological gap between a human hand and a parallel gripper and the visual gap between rendered and real images, the transferred cues remain informative, supporting their transferability across substantially different embodiments.

\subsection{Ablation Study}

\begin{table}[t]
    \vspace{5pt}
    \centering
    \caption{\textbf{Ablation study on key components.} Each design contributes to the final performance.}
    \setlength{\tabcolsep}{3pt}
    \label{tab:ablation}
    \resizebox{\linewidth}{!}{
    \begin{tabular}{l|ccc|c}
        \toprule
        \textbf{Method} & \textbf{Bottle-on-Cabinet} & \textbf{Cheese-in-Bowl} & \textbf{Bowl-on-Plate} & \textbf{Avg.} \\
        \midrule
        $K=1$ (Unimodal) 
        & $79.6 \pm 4.3$ 
        & $65.6 \pm 4.3$ 
        & $77.6 \pm 7.1$ 
        & $74.3$ \\

        w/o Contact Point 
        & $75.2 \pm 5.2$ 
        & $54.0 \pm 4.7$ 
        & $76.8 \pm 5.4$ 
        & $68.7$ \\

        w/o Sub-Goal State
        & $78.0 \pm 3.2$ 
        & $65.2 \pm 3.0$ 
        & $76.8 \pm 4.6$ 
        & $73.3$ \\

        w/o Pre-training 
        & $66.8 \pm 4.1$ 
        & $56.0 \pm 5.5$ 
        & $76.0 \pm 4.5$ 
        & $66.3$ \\

        \midrule
        \textbf{Ours} 
        & $\mathbf{80.0 \pm 6.3}$ 
        & $\mathbf{67.6 \pm 3.8}$ 
        & $\mathbf{81.6 \pm 2.6}$ 
        & $\mathbf{76.4}$ \\
        \bottomrule
    \end{tabular}
    }
\end{table}

We ablate the key design choices of Game2Policy, including game pre-training, the contact point cue, the sub-goal state cue, and the number of readout peaks, as summarized in TABLE~\ref{tab:ablation}. Ablation results show that game pre-training is critical, while the contact point and sub-goal play complementary roles. Removing game pre-training drops ACT's score under 5 demonstrations by $10.1$ points, back to $66.3$, which is essentially the Target-Only level of $65.5$, indicating that a few demonstrations alone provide insufficient supervision for learning the affordance cues. Between the two cues, the contact point contributes more: removing it costs $7.7$ points on
average, versus $3.1$ for the sub-goal. More importantly, their failure modes are complementary: losing the contact point hurts Cheese-in-Bowl most ($-13.6$), because that task's difficulty lies in grasping, while losing the sub-goal hurts Bowl-on-Plate most ($-4.8$), because its difficulty lies in placement. Each single-cue variant still outperforms the no-pretraining baseline, indicating that both cues carry independent transferable signal. Finally, collapsing the readout to a single peak ($K=1$) has a minor effect on
average success ($-2.1$ points) but a clear effect on stability: on Bowl-on-Plate the standard deviation inflates from $2.6$ to $7.1$. Because contact locations are genuinely multimodal, retaining the top-$K$ peaks primarily buys run-to-run robustness rather than higher mean performance.

\subsection{Data-Level Analysis of Collection Paradigms}

\begin{table}[t]
  \centering
  \vspace{5pt}
  \caption{\textbf{Data-level analysis.} Our method provides annotation-free labels with structural advantages over labels extracted from human video and exhibits greater behavioral diversity than simulation-generated demonstrations. }
  \label{tab:data_quality}
  \small
  \setlength{\tabcolsep}{4pt}
  \renewcommand{\arraystretch}{1.1}
  \begin{subtable}[t]{\columnwidth}
    \centering
    \caption{Label Quality vs.\ Human Video}
    \label{tab:vs_human}
    \begin{tabular}{@{}c|cc@{}}
      \toprule
      Metric & \textbf{Ours} & Human Video~\cite{damen2022epic} \\
      \midrule
      Signal Dropouts\,/s $\downarrow$ & \textbf{0.05}   & 0.48 \\
      L/R Label Flips $\downarrow$     & \textbf{0.0\%}  & 3.4\% \\
      Hand Coverage $\uparrow$         & \textbf{98.8\%} & 91.7\% \\
      Contact PCK@2 $\uparrow$         & \textbf{100\%}  & 97.5\% \\
      Manual Labels $\downarrow$       & \textbf{0}      & 89{,}977 \\
      \bottomrule
    \end{tabular}
  \end{subtable}
  \par\vspace{2ex}
  \begin{subtable}[t]{\columnwidth}
    \centering
    \caption{Behavioral Diversity vs.\ Simulation}
    \label{tab:vs_sim}
    \begin{tabular}{@{}c|ccc@{}}
      \toprule
      Metric & \textbf{Ours} & LIBERO Spatial & LIBERO Goal \\
      \midrule
      Approach Spread $\uparrow$      & \textbf{0.801} & 0.304 & 0.353 \\
      Unimodal Groups $\downarrow$    & \textbf{0/11}  & 5/10  & 5/10 \\
      Occupancy Entropy $\uparrow$    & \textbf{0.81}  & 0.47  & 0.28 \\
      Contact Chain Length $\uparrow$ & \textbf{17.7}  & 1.04  & 0.98 \\
      \bottomrule
    \end{tabular}
  \end{subtable}
\end{table}

\begin{figure}[!t] 
\raggedleft 
\includegraphics[width=0.48\textwidth]{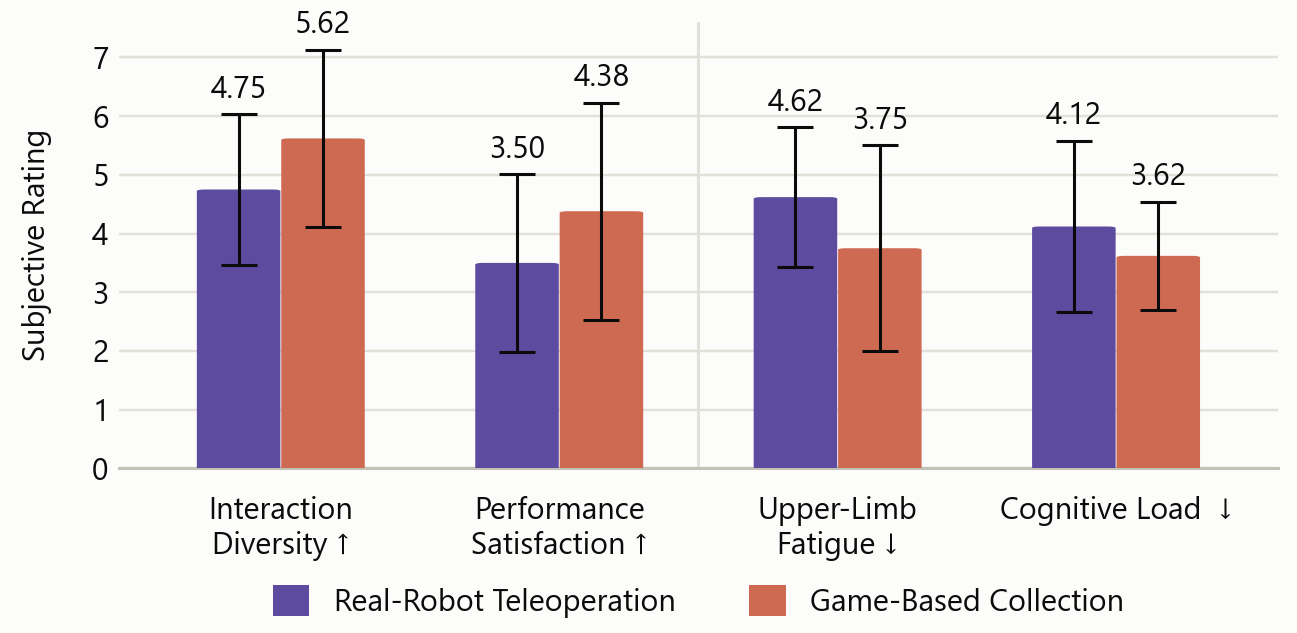}
\caption{\textbf{User study.} Gamification lowers physical and cognitive burden while sustaining engagement, offering a more sustainable foundation for crowdsourced data collection.}
\label{fig:user_study}
\end{figure}

We provide a quantitative analysis targeting the main weaknesses of existing collection paradigms, along with a user study. First, regarding label quality, our data shows structural advantages over egocentric human video data, as shown in TABLE~\ref{tab:vs_human}. Compared with EPIC-KITCHENS-100~\cite{damen2022epic}, the signal dropout rate drops from $0.48$ to $0.05$ events per second, and manual annotation requirements fall from $89{,}977$ to $0$. This gap is structural: the video paradigm must pass through a perception pipeline, which inevitably degrades under occlusion and motion blur, whereas we read hand poses and
contact events directly from the game engine state. Subtask boundaries are likewise obtained from success predicates that the game already evaluates to drive progression. The supervision thus emerges as a by-product of the game being playable and requires no additional labeling.

Second, regarding behavioral diversity, our data exhibits advantages over simulation-generated demonstrations, as shown in TABLE~\ref{tab:vs_sim}. Compared with those generated in LIBERO
Spatial and Goal, our data shows $2.3$--$2.6\times$ greater approach spread
and contact chains more than an order of magnitude longer ($17.7$ vs.\ $1.04$
and $0.98$); in each LIBERO suite, $5$ of $10$ task groups collapse onto a
single approach mode, whereas none of our $11$ groups do. Scripted generation
converges on one canonical solution per task and resets between short
episodes, whereas players pursuing a recipe choose their own approach
directions and chain dependent interactions across a long horizon.

In addition, we conduct a user study with 10 participants comparing our gamified VR platform (Project Kitchen) against teleoperation. The questionnaire comprised the SSQ, NASA‑TLX, and custom items on naturalness, attention, and behavioral diversity. We select four representative indicators for cross‑condition comparison: upper‑limb fatigue, cognitive load, behavioral diversity, and performance satisfaction. Results (Fig.~\ref{fig:user_study}) show that Project Kitchen reduces upper‑limb fatigue and cognitive load, increases behavioral diversity (consistent with our data‑level diversity metrics in Table Vb), and improves performance satisfaction relative to teleoperation. Overall, our paradigm lowers physical and cognitive burden while promoting diversity and subjective achievement, offering a more sustainable foundation for large‑scale crowdsourced data collection.






\section{Conclusion}

\begin{figure}[!t] 
\vspace{5pt}
\raggedleft 
\includegraphics[width=0.48\textwidth]{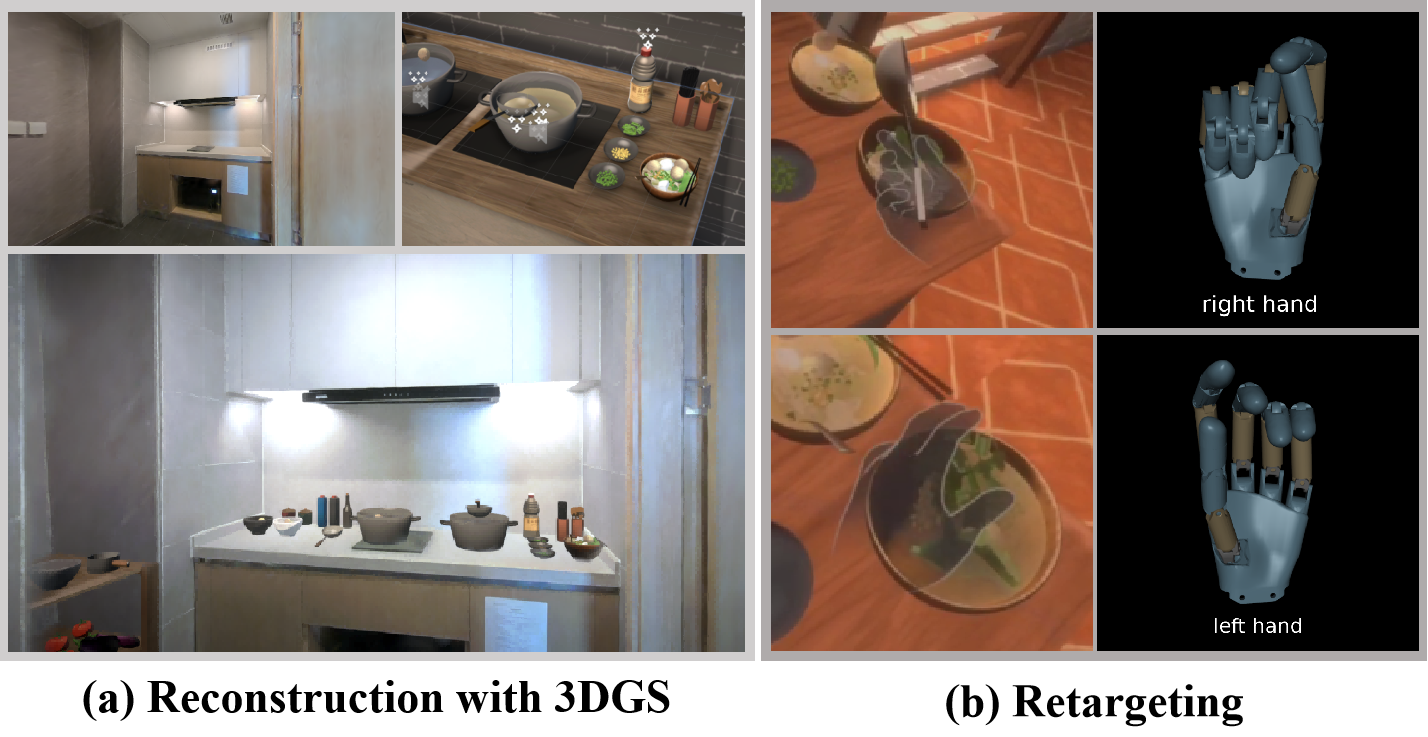}
\caption{\textbf{Future work.} Bridging the visual game-to-real gap with 3D Gaussian Splatting and retargeting gameplay hand motion to dexterous robot hands for direct policy learning.}
\label{fig:future_work}
\end{figure}

We present Project Kitchen, a gamified VR platform for hardware-decoupled data collection, and Game2Policy, which transfers embodiment-invariant affordance cues from gameplay to downstream policies via pre-training on game data and fine-tuning with few real demonstrations. In the few-shot setting, our method improves simulation success rates by 10.0 points across multiple policies and enables effective few-shot transfer on real robots, substantially reducing the need for costly teleoperated data. Our dataset offers annotation-free labels and greater behavioral diversity than simulation, while a user study indicates lower fatigue and cognitive load, together with higher performance satisfaction, than teleoperation. Overall, our framework provides a sustainable and scalable foundation for crowdsourcing manipulation data.

\textbf{Future work.}
Despite the promising results, our current framework is limited by the scale of data collection, affordance-level supervision, and the visual game-to-real gap. Future work will scale data collection through distributed deployment to broader user populations, leverage 3D Gaussian Splatting (3DGS) to reduce the visual domain gap, and retarget gameplay hand motions to dexterous robot hands for direct policy learning. We have made initial progress in the latter two directions, as shown in Fig.~\ref{fig:future_work}, toward fully leveraging large-scale gameplay data for generalist robot policy learning.




\balance

\bibliographystyle{ieeetr}
\bibliography{root}

\end{document}